\documentclass{ecai}
\usepackage{times}
\usepackage{graphicx}
\usepackage{latexsym}

\usepackage{amsmath,amsfonts,amssymb,amsthm,mathrsfs,url,array}
\usepackage{subfigure}
\usepackage[algoruled,vlined,linesnumbered]{algorithm2e}

\newtheorem{Def}{Definition} 
\newtheorem{Th}{Theorem}

\newtheorem{Lm}{Lemma}
\newtheorem{Co}{Corollary}
\newtheorem{observation}{Observation}

\allowdisplaybreaks[2]

\usepackage{color}

\newcommand{\namecite}[1]{\cite{#1}}

\newcommand{\search}{\mathit{SEARCH}}
\newcommand{\exactquery}{\mathit{ExactQuery}}
\newcommand{\approxquery}{\mathit{ApproxQuery}}
\newcommand{\xorquery}{\mathit{XorQuery}}
\newcommand{\lowerbound}{\mathit{LowerBound}}
\newcommand{\upperbound}{\mathit{UpperBound}}
\newcommand{\pointquery}{\mathit{PointQuery}}
\newcommand{\neighborquery}{\mathit{NeighborQuery}}

\begin{document}

\title{Towards Efficient Discrete Integration\\
via Adaptive Quantile Queries}

\author{
  Fan Ding\institute{Purdue University, USA, email: \{ding274, yexiang\}@purdue.edu}
  \and
  Hanjing Wang\institute{Shanghai Jiao Tong University, China, email: wanghanjingwhj@sjtu.edu.cn}
  \and
  Ashish Sabharwal\institute{Allen Institute for AI, USA, email: ashishs@allenai.org}
  \and
  Yexiang Xue$^1$
}

\maketitle

\begin{abstract}
  Discrete integration in a high dimensional space of $n$ variables poses fundamental challenges. The WISH algorithm reduces the intractable discrete integration problem into $n$ optimization queries subject to randomized constraints, obtaining a constant approximation guarantee. The optimization queries are expensive, which limits the applicability of WISH. We propose AdaWISH, which is able to obtain the same guarantee, but accesses only a small subset of queries of WISH. For example, when the number of function values is bounded by a constant, AdaWISH issues only $O(\log n)$ queries. The key idea is to query adaptively, taking advantage of the shape of the weight function being integrated. In general, we prove that AdaWISH has a regret of only $O(\log n)$ relative to an idealistic oracle that issues queries at data-dependent optimal points. Experimentally, AdaWISH gives precise estimates for discrete integration problems, of the same quality as that of WISH and better than several competing approaches, on a variety of probabilistic inference benchmarks. At the same time, it saves substantially on the number of optimization queries compared to WISH. On a suite of UAI inference challenge benchmarks, it saves $81.5\%$ of WISH queries while retaining the quality of results.
\end{abstract}

\section{INTRODUCTION}



\begin{figure}[t]
	\centering
	\includegraphics[width=.8\linewidth]{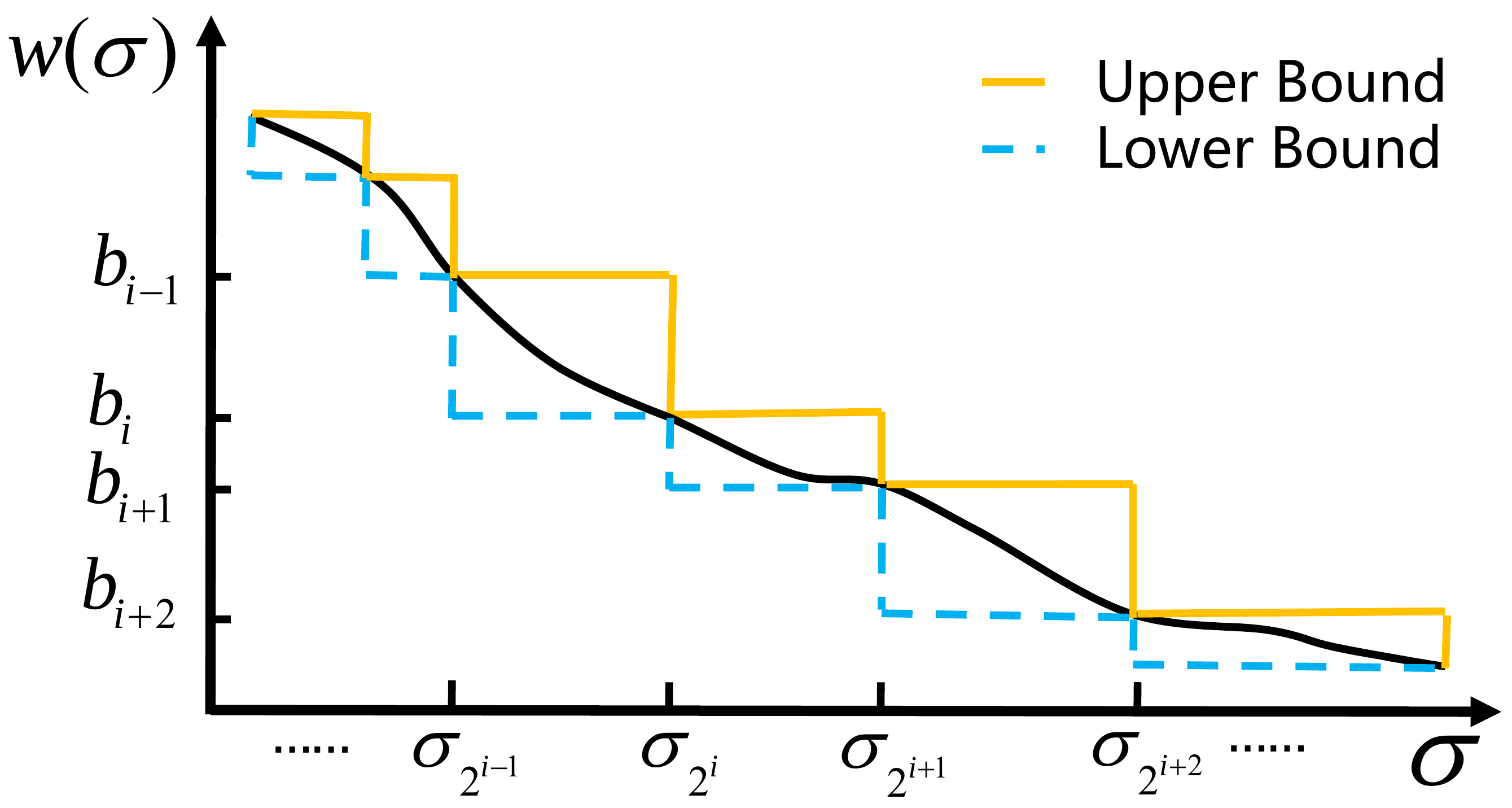}
	\includegraphics[width=.8\linewidth]{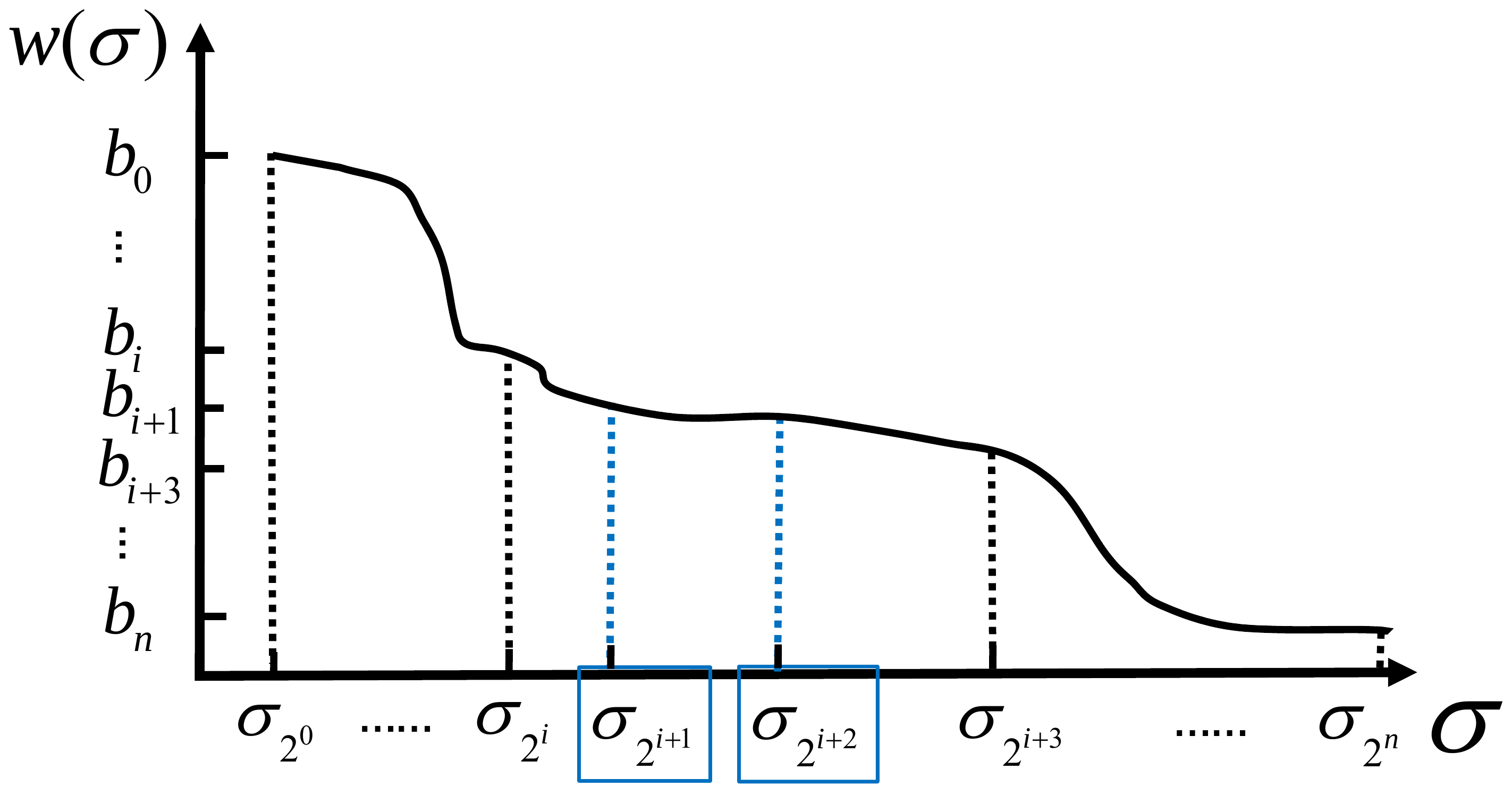}
	\caption{\textbf{(Top)} WISH obtains a constant approximation of $W=\sum{w(\sigma)}$
	by querying quantiles of $w$ that are exponentially apart, namely, $b_i$, the $2^i$-th largest item for $w$. The solid and dashed curves show the corresponding upper and lower bounds after knowing the values of these quantiles, which form a 2-approximation. \textbf{(Bottom)} AdaWISH makes queries adaptively. The set of queried points forms a subset of those of WISH. In this example, AdaWISH does not need to query the two points marked with squares, since their values can be approximately inferred from their left and right neighbors.}
	\label{fig:slices}
	\vspace{-0.1in}
\end{figure}

Discrete integration in a high dimensional space poses fundamental challenges in
scientific computing. Yet, it has numerous applications in artificial intelligence,
machine learning, statistics, biology, and physics~\cite{aziz2015stable,vlasselaer2016knowledge}. 
In probabilistic inference, discrete integration is crucial for computing core quantities such as the partition function and marginal probabilities of probabilistic graphical models.
The key challenge is the exponential growth in the volume of the space as the dimensionality increases, commonly known as the curse of dimensionality.


A fruitful line of work, based on hashing and optimization,
is able to achieve constant-factor upper and lower bounds to the discrete integration problem \cite{Ermon13Wish,belle2015hashing,asterisldpc,Gomes06XORCounting,Gomes2006Sampling,kuck2019adaptive,soos2019bird}.
The key idea is to transform the discrete integration problem into optimization problems 
subject to additional randomly sampled parity constraints.
Each imposed parity constraint randomly cuts the original space by half. If there is one element of interest in a subspace formed by applying $k$ random parity constraints, then there should be approximately $2^k$ elements of interest in the original space. 

The WISH algorithm of Ermon et al.~\namecite{Ermon13Wish} gave the first constant-factor approximation guarantee for integrating a weight function.
%
They followed a two-step approach.
In the first step, they worked out a logarithmic slicing schedule, which was able to
obtain a constant approximation guarantee by querying only the quantiles of the 
weight function that are exponentially apart; i.e., the $2^i$-th largest element
$b_i$ of the weight function, for various values of $i$. 
See Figure \ref{fig:slices} (Top) for an illustration.
The second step was to obtain $b_i$ using hashing and optimization. They completed this step by querying optimization oracles subject to randomized parity constraints. 

We propose AdaWISH, which is able to obtain the same constant approximation 
guarantees for weighted integration, but with a significantly fewer number of queries to the optimization oracle. 
The key idea is to make quantile queries in an \emph{adaptive} way, using divide-and-conquer. 
See Figure~\ref{fig:slices} (Bottom) for the intuitive idea. Here, WISH queries all quantiles. AdaWISH, on the other hand, skips querying the two quantiles shown in boxes
because the neighboring black quantiles to their left and the right, which 
sandwich the quantiles, are close in value.
Exploiting this observation, one can prove, for instance, that AdaWISH needs only $O(\log n)$ queries if the function has a fixed number of outputs. This case coincides
with the results of 
\namecite{Chakraborty2016ApproxCount2}. While their work focused on the unweighted case, our results apply to a more general weighted case.

We analyze the performance of AdaWISH when it accesses 
 oracles in two different approximate ways. 
In the first way, when queried for one specific quantile $b_i$, the oracle returns a value that is guaranteed to be bounded between neighboring quantiles $b_{i \pm c}$ with a high probability. 
This setting corresponds precisely to the probabilistic 
bound one obtains with hashing and optimization. 
In the second way, we consider another natural setting where the oracle returns a pointwise approximation, i.e., a value within a known constant factor of $b_i$. 
%
%
%
In both settings, we are able to prove the following: (i) \textbf{\textit{(Upper bound)}} AdaWISH makes at most $n$ queries, forming a subset of WISH's queries while achieving an identical constant approximation guarantee; 
(ii) \textbf{\textit{(Regret bound)}}
The number of queries of AdaWISH cannot exceed a logarithmic factor times that of an ``optimal'' algorithm, which makes the least amount of queries for the same guarantee, while knowing the exact shape of the weight function a-priori;
(iii) \textbf{\textit{(Lower bound)}}  Any algorithm that guarantees a constant-approximation for all weight functions must make at least $\Omega(n)$ accesses to the oracle.
We thus conclude that the number of optimization queries made by 
AdaWISH is close to optimal. 
%
Our theoretical bounds apply when every approximate oracle query issued by WISH is solved to optimality. When some queries are not solved to optimality, WISH continues to provides strong empirical estimates of practical value (without a guarantee). AdaWISH continues to match the quality of these estimates with substantially fewer queries, thereby providing added value even in this case.

Experimentally, we test the efficacy of AdaWISH in the context of computing the partition function of random clique-structured Ising models, grid Ising models, and instances from the UAI inference competitions. Our results demonstrate that AdaWISH provides precise estimates of discrete integration problems, of the same quality as that of WISH and better than competing approaches like belief propagation, mean field, dynamic importance sampling, HAK, etc. Meanwhile, AdaWISH saves a vast majority of optimization oracle queries compared to WISH.
For example, it reduces the number of queries by 47\%-60\% on Ising models and a median of 81.5\% queries for benchmarks from the UAI inference challenge.

~\\
\noindent\textbf{Related work.} ~Over the years, many exact and approximate approaches have been proposed to tackle the discrete integration problem. 
Variational methods \cite{Jordan1999Variational,Wainwright2008} search for a tractable variational form to approximate the otherwise intractable integration. 
These methods are fast but often cannot provide tight bounds on the quality of the outcome. 
Approximate sampling techniques  \cite{Jerrum:1996:MCM:241938.241950,madras2002lectures,Poon2006mcsat} are popular, but the number of samples required to obtain a reliable estimate often grows exponentially in the problem size. 
Importance sampling approaches such as SampleSearch \cite{GOGATE2012AndOr,lou2019interleave} and methods based on knowledge compilation \cite{ChoiKisaDarwiche13} have also achieved a fundamental breakthrough in performance \cite{hazan2012partition,liu2015estimating,lou2017dynamic,friedman2018approximate}.

There have been recent developments on the use of short parity constraints to speed up computation in WISH style methods~\cite{Ermon14LowDensityParity,asterisldpc,achlioptas2017probabilistic,achlioptas2018fast}, and on a dual variant of WISH called SWITCH \cite{de2019dual}.
While these approaches reduce the empirical complexity of answering individual optimization oracle queries, our method reduces the number of queries itself, complementing these approaches.

Only a handful of hashing based approaches
provide guarantees for weighted functions \cite{Chakraborty2015WeightedCounting,ermon2013embed}.
Using the \textit{tilt} parameter of a weighted Boolean formula 
is an alternative~\cite{Chakraborty2014tiltWeightedSampling} that allows relying only on
NP oracles, which can be more efficient in practice than the MAP oracles used by WISH and AdaWISH. However, in typical applications and benchmarks including UAI instances, \textit{tilt} can be impractically large.

The adaptive approach of AdaWISH is motivated by the work of \namecite{Sabharwal18PR}, who addressed the problem of reducing the number of human annotations needed
to reliably estimate the precision-recall curve of massive noisy datasets. 
%
Our application domain, namely discrete integration, differs in a number of aspects, such as our space being exponential and thus too large to enumerate (they operated on an explicitly stored dataset), our query oracles being NP-hard (their query was human annotation of a data point), and our weight function not being guaranteed to not increase/decrease too fast (PR curves, on the other hand, must necessarily be relatively stable for large enough ranks). These differences necessitate new proof techniques.

\section{PRELIMINARIES}
\begin{algorithm}[t]
\caption{$\xorquery(i, \Sigma, w, T)$} 
\label{alg:queryXOR}  
\If{the $2^i$-th largest item $\tilde{b}_i$ has been queried before}
    {return $\tilde{b}_i$\;}
\Else{
    \For{$t=1,...,T$}{
        {Sample hash function $H_i^t$ from $\mathcal{H}_i$ and $d \in\{0,1\}^i$ uniformly at random}\\
        {$w_i^t\leftarrow \max_{\sigma}w(\sigma)$ subject to $H_i^t(\sigma) = d$}
    }
    {$M\leftarrow$ Median($w_i^1,...,w_i^T$)}\\
    {\Return $M$ as the estimate $\tilde{b}_i$}\;
}
\end{algorithm}

We consider a weighted model with $n$ binary variables $x_1,\ldots, x_n$ where $x_i\in \{0,1\}$. $x = (x_1, \ldots, x_n)^T$ is a vector which takes values from the space $\Sigma = \{0, 1\}^n$. Define a weighted function $w:\Sigma\rightarrow\mathbb{R}^+$ that assigns a non-negative weight to each element $\sigma$ in $\Sigma$. The discrete integration problem is to compute the total weight: 
$W=\sum_{\sigma\in\Sigma}w(\sigma)$.

The discrete integration problem is an important but challenging problem in machine learning. 
A recently proposed method  \textbf{W}eight-\textbf{I}ntegral-And-\textbf{S}um-By-\textbf{H}ashing (WISH) \cite{Ermon13Wish} provides a \textit{constant-approximation guarantee} to this problem. 
A careful analysis of WISH reveals that the constant approximation guarantee is achieved via two main ingredients: (i) a logarithmic slicing schedule; (ii) counting via hashing and optimization. 


~\\
\noindent\textbf{Logarithmic slicing schedule.} ~We fix an ordering for all $\sigma \in \Sigma$ such that $w(\sigma_j)\geq w(\sigma_{j+1})$ holds for all $j$, $1\leq j \leq 2^n-1$ and let $b_i$ be the $2^i$-th largest element, i.e., $b_i = w(\sigma_{2^i})$. 
The first contribution of \cite{Ermon13Wish} is a constant approximation
scheme constructed from knowing only $n+1$ points of function $w$, namely, $b_0, \ldots, b_n$. 
\begin{Lm}[Ermon et al. \cite{Ermon13Wish}]
\label{Lm:wish}
$LB = b_0 + \sum_{i=1}^n b_i (2^i - 2^{i-1})$ is a lower bound and a 2-approximation to $W$. In other words, $$LB \leq W \leq 2LB.$$
\end{Lm}
%



\noindent\textbf{Counting via hashing and optimization.} ~In order to estimate $W$, the next step is to estimate $b_i = w(\sigma_{2^i})$ for $i=0, \ldots, n$. The work of \cite{Ermon13Wish} translates 
this problem into optimization problems subject to randomized hashing (parity) constraints. 
The high-level idea is as follows. To compute $b_m$, the $2^m$-th largest item, consider $2^m$ buckets, each of which is labeled with one vector from $\{0,1\}^m$.
Suppose we hash all elements in $\Sigma$ uniformly at random into these $2^m$ buckets. Then we use an optimization oracle to compute the largest item in a given bucket. 
Because elements are hashed randomly, if we repeat this process multiple times and consistently find that the largest item in one bucket is larger than $w^*$, then we 
can conclude that there must be more than $2^m$ items larger than $w^*$, hence $b_m \geq w^*$. 
Following the same argument, if there are more than $2^m$ items 
larger than $\hat{w} > w^*$, then the optimization oracle 
should return $\hat{w}$, larger than  $w^*$. 
Combining these two points, if we repeat this experiment multiple times, the median value of the largest item in the repeated bucket experiments should reflect the actual value of
 $b_m$. 
In practice, we form the buckets with pairwise independent hashing functions:
\begin{Def}
A function family $\mathcal{H}_m=\{H_m:\{0,1\}^n\rightarrow\{0,1\}^m\}$  is \emph{pairwise independent} if the following conditions hold when $H_m$ is chosen uniformly at random from $\mathcal{H}_m$. 1) $\forall x\in\{0,1\}^n$, the random variable $H_m(x)$ is uniformly distributed in $\{0,1\}^m$. 2) $\forall x_1,x_2\in\{0,1\}^n$ and $x_1\neq x_2$, random variables $H_m(x_1)$ and $H_m(x_2)$ are independent.
\end{Def}
For one configuration $\sigma \in \Sigma$, we say $\sigma$ is hashed to the bucket labeled with $d\in \{0,1\}^m$ by function $H_m$ if $H_m(\sigma)=d$. 
In practice, pairwise independent hash functions are constructed with random
parity functions. Let matrix $A\in\{0,1\}^{m\times n}$ be a randomly sampled
0-1 matrix. One can prove that function family $\{h_A(x) = Ax \mbox{ mod } 2\}$ is pairwise independent. 
$\xorquery(i, \Sigma, w, T)$ (Algorithm \ref{alg:queryXOR}) demonstrates the actual implementation to compute the $2^i$-th largest item $b_i$. 
The formal mathematical result in \cite{Ermon13Wish} bounds the returned
value $M$ of the algorithm between $b_{\min\{i+c,n\}}$ and $b_{\max\{i-c,0\}}$ for any $c \geq 2$, a small range around $b_i$:
\begin{Lm}\label{Lm:xorcounting}
    [Ermon et al. \cite{Ermon13Wish}] Let $M$ be the value returned by 
    $\xorquery(i, \Sigma, w, T)$. Then for any $c \geq 2$, there exists an
    $\alpha^*(c) > 0$ such that for $0 < \alpha < \alpha^*(c)$, 
    \begin{equation*}
        Pr\left(M \in [b_{\min\{i+c,n\}},b_{\max\{i-c,0\}}]\right) \geq 1 - \exp(-\alpha T)
    \end{equation*}
\end{Lm}
Combining Lemma~\ref{Lm:xorcounting} with the logarithmic slicing schedule,
the authors of \cite{Ermon13Wish} are able to provide a constant approximation algorithm for the discrete integration problem with at most a logarithmic number of accesses to the optimization queries:
\begin{Th}
    [Ermon et al. \cite{Ermon13Wish}] For $\delta>0$, WISH algorithm makes $\Theta(n \log n \log(1/ \delta))$ MAP queries and with probability at least $1-\delta$, outputs a 16-approximation of $W$.
\end{Th}
  


\section{AdaWISH: ADAPTIVE DISCRETE INTEGRATION}
In WISH, we query all the $n+1$ quantiles $b_0, b_1, \ldots, b_n$. In practice, 
the number of queries can be reduced due to the shape of the $w$ function. 
For example, the two quantiles in Figure~\ref{fig:slices} (Bottom)
are sandwiched between the left and right quantile in black, which are 
close in values. 
From this observation, we do not need to query the two blue quantiles 
and instead can use the values of the neighboring quantiles to 
replace their values. 
%


Motivated by this example, we propose \textbf{Ada}ptive-\textbf{W}eight-\textbf{I}ntegral-And-\textbf{S}um-By-\textbf{H}ashing (AdaWISH), an algorithm which makes queries adaptively using divide-and-conquer. 
%
%
The detailed AdaWISH algorithm is shown in Algorithm \ref{alg:AdaWISHXOR}, where the algorithm starts with estimating
the quantiles $b_0, \ldots, b_n$ in the entire range ($\search(\Sigma, w, \beta, 0, n)$), and recursively breaks the range by the middle point (geometric mean) using divide-and-conquer (shown as $\search(\Sigma, w, \beta, l, m)$ and $\search(\Sigma, w, \beta, m, r)$) until the stopping condition is met. 
In this algorithm, $\beta > 1$ is a user-defined parameter for trade-off. The larger $\beta$ is, the 
worse the approximation guarantee, but the fewer number of queries
that AdaWISH has to make.  
AdaWISH also depends on three query functions, $\approxquery$, 
$\lowerbound$, and $\upperbound$, which are implemented
differently assuming different types of oracles, which will
be introduced shortly. 
The detailed implementation of these functions will be 
discussed together with the specific oracle. 
Here, at a high level, $\approxquery(i, \Sigma, w)$ outputs a point estimate of quantile $b_i$, while $\lowerbound(i, \Sigma, w)$ ($\upperbound(i, \Sigma, w)$) outputs a lower (upper) bound of the quantile $b_i$, respectively. 
In $\search(\Sigma, w, \beta, l, r)$, there are two stopping conditions once we have queried the left point $\tilde{b}_l$ and right point $\tilde{b}_r$. 
The first is $r=l+1$, in which there is no point to query between 
the two quantiles. The second condition is that the values of $\tilde{b}_l$ and $\tilde{b}_r$ 
are close: $\tilde{b}_l\leq\beta \tilde{b}_r$. This suggests that the $w$ function stays roughly flat between $b_l$ and $b_r$. 
If one of these two conditions is met, we stop and replace all the $b_i$ between
$\tilde{b}_l$ and $\tilde{b}_r$ with the value of $\tilde{b}_r$. Otherwise, we break the range $l\ldots r$ through the middle point  $m=\lfloor \frac{r+l}{2}\rfloor$ then recursively calls $\search$ on two sub-partitions.



%
We analyze the performance of AdaWISH, when it accesses 
quantile oracles in two different ways. 
In the first way, when queried with one quantile,  the oracle returns a value that is guaranteed to be bounded between neighboring quantiles with high probability. 
This type of oracle corresponds the case of using hashing and randomization, i.e., using $\xorquery$ in algorithm~\ref{alg:queryXOR}.
In another way, we consider a natural case where every oracle access returns a quantile in its point-wise bound, e.g., the multiplicative distance between the returned and the exact values are within a constant range. 

\subsection{Neighboring query oracle}

We focus mainly on the $\neighborquery$ oracle. When we query a particular quantile $b_i$, $\neighborquery$ returns a value that is guaranteed to be bounded between neighboring quantiles with high probability. To be precise, given $c \geq 2$ and $\delta > 0$, for all $i$, $\neighborquery(i, \Sigma, w)$ returns an estimation of $b_i$ that is guaranteed to be in the range $[b_{\max{\{i-c,0\}}}, b_{\min{\{i+c,n\}}}]$ with probability at least $1 - \delta$. 
Notice that we can use hashing and optimization to build $\neighborquery$ oracles. 
According to Lemma~\ref{Lm:xorcounting}, if we set $T = \lceil\frac{\ln(1/\delta)}{\alpha(c)}\ln{n}\rceil$, then 
the output of $\xorquery(i, \Sigma, \beta, w, T)$  (Algorithm \ref{alg:queryXOR}) satisfies the conditions of a $\neighborquery$ oracle. 


%

We implement the AdaWISH algorithm 
as follows: 
for all $i$, $\approxquery(i,\Sigma,w)$ returns exactly $\neighborquery(i,\Sigma,w)$. 
$\lowerbound(i,\Sigma,w)$ returns the value of $\neighborquery(\min\{i+c,n\},\Sigma,w)$, which with high probability is a lower bound for $b_i$ based on Lemma~\ref{Lm:xorcounting}. 
$\upperbound(i,\Sigma,w)$ returns $\neighborquery(\max\{i-c,0\},\Sigma,w)$, which with high probability is an upper bound for $b_i$. 
We implement a look-up table within $\neighborquery$. If $\neighborquery(i,\Sigma,w)$ is called
twice with the same parameters, then the second time $\neighborquery$ directly returns the result 
without re-computing. 
We can prove that the AdaWISH algorithm implemented in this way gives a constant factor approximation:  

\begin{Th}\label{th:2beta2c}
\textbf{(Constant Approximation)} Let $w,\Sigma,\delta,$ and $c \geq 2$ be as defined earlier. For any $\kappa > 2^{2c}$, the output of AdaWISH (Algorithm \ref{alg:AdaWISHXOR}) on input $(\Sigma, w, \kappa / (2^{2c}))$, assuming oracles with neighboring bound, is a $\kappa$-approximation of $W$ with probability $1-\delta$.\label{Th:constant}
\end{Th}

At a high level, the two stopping conditions guarantees the constant approximation guarantee
of Theorem~\ref{th:2beta2c}. The first stopping condition is $r=l+1$, in which AdaWISH queries
all the quantiles within the range from $l$ to $r$. The approximation guarantee can be enforced 
with the same argument in the proof of WISH. The second stopping condition is $\tilde{b_l} \leq \beta \tilde{b_r}$, which suggests that function $w$ decreases very slowly in the range from $l$ to $r$. In 
this case, we can approximate the values of $w$ in between with the queried value from either end, without losing much accuracy. 
\begin{proof} (Theorem~\ref{Th:constant})
The approximation $\tilde{W}$ given by AdaWISH is $\tilde{W} = \tilde{b}_0+\sum_{i=0}^{n-1}\tilde{b}_{i} 2^i$. 
Define $L'=b_0+\sum_{i=0}^{n-1}b_{\min\{i+c,n\}}2^i$ 
and $U'=b_0+\sum_{i=0}^{n-1}b_{\max\{i-c,0\}}2^i$. 
From Lemma 2 in \cite{Ermon13Wish}, we have $L' \leq W \leq U'$ (1) and $U' \leq2^{2c}L'$ (2). 
We are going to prove $\tilde{W}$ satisfies $L' / \beta \leq \tilde{W} \leq \beta U'$ (3) with high probability. 
Combining (2) and (3), we have $L' / \beta \leq \tilde{W} \leq \beta 2^{2c} L'$ (4). 
Combining (1) (2) and (4), we can have 
$W \leq U' \leq 2^{2c}L' \leq 2^{2c} \beta \tilde{W}$. In short, $W \leq 2^{2c} \beta \tilde{W}$ (5). 
Combining (1) and (4), we have $\tilde{W} \leq \beta 2^{2c} L' \leq \beta 2^{2c} W$ (6). 
Together (5) and (6) imply that $\tilde{W}$ is a $\beta 2^{2c}$-approximation. 
Under the condition of this theorem,  $\beta$ is set to $\kappa/2^{2c}$. Therefore,
the overall approximation factor is $\kappa$. 

We are left to prove (3): $L' / \beta \leq \tilde{W} \leq \beta U'$. Comparing
the corresponding terms of $L'$, $\tilde{W}$, and $U'$, it is sufficient to prove that for all $i$, 
the following Inequality (7) holds:
$$ b_{\min\{i+c,n\}} / {\beta}\leq\tilde{b}_i\leq\beta {b_{\max\{i-c,0\}}}.$$
Consider the two stopping conditions for AdaWISH. 
If stopping condition 1 is met, $\tilde{b}_l$ and $\tilde{b}_r$ are from $\approxquery$, hence 
(7) holds because of Lemma~\ref{Lm:xorcounting}. 
If stopping condition 2 is met for a range from $l$ to $r$, 
we have $\tilde{b}_l \leq \beta \tilde{b}_r$ (8) 
Here, $\tilde{b}_l$ is the result from $\xorquery(\max\{l-c, 0\}, \Sigma, w, T)$.
According to Lemma~\ref{Lm:xorcounting}, $\tilde{b}_l \geq b_l$ (9) with high probability.
For the same reason, $\tilde{b}_r \leq b_r$ (10) with high probability. 
The estimation of $\tilde{b}_i$ in between are replaced with $\tilde{b}_r$. 
Hence, $\tilde{b}_i = \tilde{b}_r \leq b_r \leq b_{\max\{i-c,0\}} \leq \beta b_{\max\{i-c,0\}}$ (11).
Here, the first inequality is due to (10), the second due to monoticity of the quantiles. 
Similarly, we have
$\tilde{b}_i = \tilde{b}_r \geq \tilde{b}_l / \beta \geq b_l / \beta \geq b_{\min\{i+c, n\}} / \beta$ (12),
where the first inequality is due to the stopping condition. 
With (11, 12), we also get (7). 
\end{proof}

To understand AdaWISH in terms of the number of calls to $\neighborquery$, we analyze the upper bound, regret bound and asymptotic lower bound respectively. 
%

\begin{Th}\label{Th:n_of_call}
\textbf{(Upper Bound)} Under the conditions of Theorem \ref{th:2beta2c}, the number of $\neighborquery$ calls is at most $n+1$, which is only a subset of that of WISH.
\end{Th}
To prove Theorem \ref{Th:n_of_call}, in the worst case AdaWISH has to query all $b_0, \ldots, b_n$, which is exact the case of WISH. The lookup table implemented in the oracle guarantees the same query will not be computed twice. 
In practice, AdaWISH can save a lot of queries. For example: 
\begin{observation}
If function $w(\sigma)$ only has a fixed set of $k$ different values, then AdaWISH makes only $O(\log n)$ accesses to $\neighborquery$.
\label{obs:fixedvalues}
\end{observation}

Intuitively, since function $w(\sigma)$ only has a fixed set of $k$ values, it requires AdaWISH to search for the $k-1$ quantiles, where the function values change from one value to another. AdaWISH uses binary search. Therefore, it requires $O(\log n)$ queries to determine one point. Since $k$ is a constant, the total number of queries is  $O(\log n)$ to determine the entire $w$ function. 
~\\
\noindent\textbf{Regret bound.} ~We consider an ``optimal'' algorithm, which is guaranteed to produce a $\kappa$-approximation by issuing the least number of accesses to an $\exactquery$ oracle. 
When queried on a particular $b_i$, $\exactquery$ returns the exact value of $b_i$. 
%
We allow the optimal algorithm to know the shape of the $w$ function a priori.
The optimal algorithm issues $\exactquery$ very smartly. 
Let $B = \{\tilde{b}_0, \tilde{b}_{i_1}, \ldots, \tilde{b}_{i_k}, \tilde{b}_n\}$ be the set of points the optimal algorithm issues $\exactquery$ on. We can bound the sum $W$ between the upper bound $UB = b_0 +  b_0 (2^{i_1} - 2^0) + \sum_{l=1}^{k-1} b_{i_l} (2^{i_{l+1}} - 2^{i_l}) + b_{i_k} (2^n - 2^{i_k})$ 
and the lower bound 
$LB = b_0 + b_{i_1}(2^{i_1} - 2^0) + \sum_{l=2}^k b_{i_l} (2^{i_{l}} - 2^{i_{l-1}}) + b_n (2^n - 2^{i_k})$.
%
We require the optimal algorithm to obtain a $\kappa$-approximation, i.e., $LB \leq UB \leq \kappa LB$ must hold. 
The \textbf{\textit{mathematical definition}} of the optimal algorithm is the one that minimizes the
size of $B$, i.e., the set of queried points, while enforcing $LB \leq UB \leq \kappa LB$. 
We call the number of accesses to $\exactquery$ of this optimal algorithm, i.e., the size of $B$, $OPT$. 
We compare the number of QUERY accesses of AdaWISH against $OPT$ and show the difference is within a multiplicative $O(\log n)$ factor: 
\begin{algorithm}[t]
\caption {AdaWISH($\Sigma, w, \beta$)} 
\label{alg:AdaWISHXOR}
{$n=\log_2|\Sigma|$}\;
{$\tilde{b}_0,\ldots.,\tilde{b}_n\leftarrow \search(\Sigma, w, \beta, 0, n)$}\;
{$\tilde{W}\leftarrow \tilde{b}_0+\sum_{i=0}^{n-1}2^i\tilde{b}_{i}$}\;
\Return $\tilde{W}$
\end {algorithm}

\begin {algorithm}[t]
\DontPrintSemicolon
\caption {$\search(\Sigma, w, \beta, l, r)$}
\label{alg:oracleXOR}  
\If{$r==l+1$}{
    //\text{\ stopping condition 1 met}\;
    {$\tilde{b}_l\leftarrow \approxquery(l, \Sigma, w)$}\;
    {$\tilde{b}_r\leftarrow \approxquery(r, \Sigma, w)$}\;
}
\Else{
    {$\tilde{b}_{l}\leftarrow \upperbound(l, \Sigma, w)$}\;
    {$\tilde{b}_{r}\leftarrow \lowerbound(r, \Sigma, w)$}\;
    \If {$\tilde{b}_{l}\leq \beta\tilde{b}_{r}$}{
    //\text{\ stopping condition 2 is met}\;
        \lFor {$i\in\{l,\ldots,r-1\}$}{
            {$\tilde{b}_i \leftarrow \tilde{b}_{r}$}
        }
    }
    \Else{{$m\leftarrow\lfloor \frac{r+l}{2}\rfloor$} //\text{\ bisect the interval}\;
    {$\tilde{b}_l,\ldots,\tilde{b}_m\leftarrow \search(\Sigma, w, \beta, l, m)$}\;
    {$\tilde{b}_{m},\ldots,\tilde{b}_r\leftarrow \search(\Sigma, w, \beta, m, r)$}
    }
}
{\Return $\tilde{b}_l,...,\tilde{b}_r$}
\end {algorithm}
\begin{Th}\label{Th:regret_b}
\textbf{(Regret Bound)} Suppose $\kappa=2\beta\gamma^2$, AdaWISH in algorithm \ref{alg:AdaWISHXOR} on input $(\Sigma, w, \beta)$, assuming neighboring query oracle, calls $\neighborquery$ at most $(OPT-1)(2+\log_2n)+1$ times.
\end{Th}
This says that the number of QUERY calls made by AdaWISH is roughly $O(OPT\cdot\log_2n)$. 
Theorem~\ref{Th:regret_b} is a strong result in the following sense: first, the ``optimal'' algorithm has access to the $\exactquery$ oracle, while AdaWISH only accesses $\neighborquery$ which returns approximations to $b_i$'s. Even with approximate oracles, the number of queries AdaWISH needs is within a logarithmic factor of that of WISH. Second, the ``optimal'' algorithm does not need to follow the query schedule of WISH or AdaWISH. We give the algorithm the privilege of accessing $\exactquery$ oracles. It is at the ``optimal'' algorithm's discretion how to make use of such a privilege. 
The high level idea to prove Theorem \ref{Th:regret_b} is as follows. Suppose $q_1,...,q_{OPT}$ are the actual query points of the optimal algorithm. Because AdaWISH uses a binary search, i.e., it always almost splits an interval at its geometrical middle point. Then it takes AdaWISH roughly $O(\log_2n)$ splits to ``locate'' one query point $q_i$ of the optimal algorithm (more precisely, find a point that is sufficiently close to $q_i$ that guarantees the approximation bound). Hence, the total number of queries of AdaWISH is bounded by OPT times $\log_2n$. Our accurate proof to Theorem \ref{Th:regret_b} is based on walking through the actual calling map of the function $\search$, where each node in this map represents an actual interval
that $\search$ is called. 

\begin{proof} (Theorem \ref{Th:regret_b})
Suppose the optimal algorithm calls $\exactquery$ at points $q_1,...,q_{OPT}$ as above. These points split the entire range between $[\sigma_{2^0},\sigma_{2^n}]$ into OPT-1
segments. We call one segment of this type an opt-segment. We also define the following tuple $\langle \sigma_{2^l},\sigma_{2^r},d,\mathrm{SMALL/BIG}\rangle$, where ($\sigma_{2^l},\sigma_{2^r}$) is an interval on which the function SEARCH is called during the execution of AdaWISH, d is the depth of this recursive call, and the forth entry is either SMALL or BIG. It is BIG if and only if the interval $(\sigma_{2^l},\sigma_{2^r})$ covers at least one complete opt-segment. The log distance of a tuple is measured as $r-l$. Two tuples are treated as cousins if they are called by a single ORACLE function as two children calls(i.e.,one is $(\sigma_{2^l},\sigma_{2^m})$, while the other is ($\sigma_{2^m},\sigma_{2^r}$)). We can merge two cousin tuples. The result of merging is the tuple representing the parent function that called the functions represented by these two cousin tuples. Notice that the log distance of the merged tuple is between 2 times of those of the two cousin tuples.

We start with the set $\Phi$, made of tuples $\langle \sigma_{2^l},\sigma_{2^r},d,\mathrm{SMALL/BIG}\rangle$ such that $(\sigma_{2^l},\sigma_{2^r})$ enters the stopping condition of function SEARCH during the execution of the algorithm. We repeat the following merge operation on tuples in $\Phi$ until no tuple in $\Phi$ is tagged with SMALL:

\begin{enumerate}
    \item Choose a tuple from $\Phi$ that has the largest depth d among those tagged with SMALL.
    \item Merge this tuple with its cousin tuple.
    \item Add the merged tuple back into $\Phi$.
\end{enumerate}

\noindent We refer to the set $\Phi$ obtained after merging all SMALL tuples as $\Phi^E$. First, all tuples in $\Phi^E$ are tagged BIG. Therefore, each of them contains an opt-segment. Second, since the tuples in $\Phi^E$ are still from the SEARCH function, these tuples must be non-overlapping. Based on these two observations, the number of tuples in $\Phi^E$ is bounded by the number of opt-segment, which is OPT-1. We should notice that the merge operations that each tuple in $\Phi^E$ have gone through must be bounded by $O(\log_2n)$, since the largest possible log distance is no more than $\log_22^{n}=n$ (the entire range), and each merge operation nearly doubles the log distance. Considering that there are some interval whose log distance is odd, we need to perform at most $\lceil\log_2n\rceil \leq \log_2n+1$ merge operations.

Now we count the number of QUERY calls that AdaWISH makes. It is easy to see that
\begin{equation}
\#QUERY=|\Phi^E|+\#MERGE+1
\end{equation}
where $\#$MERGE is the total number of merge operations performed. This number is bounded by
\begin{equation}
|\Phi^E|+|\Phi^E|\cdot max\{\#\text{MERGE for one tuple in }\Phi^E\}+1
\end{equation}
Combined with the fact that $|\Phi^E|\leq OPT-1$ and $\#MERGE$ for one tuple in $\Phi^E\leq \log_2n+1$, we obtain the claimed bound.
\end{proof}


\noindent\textbf{Asymptotic lower bound.} ~What is the minimum number of calls to the oracle in order to guarantee a constant factor approximation to $W$, let's say, a $\kappa$-approximation? 
In this section, assuming $ExactQuery(i)$ returns the exact value of $b_i$, we prove that at least $(1-\epsilon)\frac{n}{\kappa^2}$ accesses to $ExactQuery$ is needed to obtain a $\kappa$-approximation in the worst case, even for the optimal algorithm, which knows the shape of the $w$ function a-priori (Corollary~\ref{Co:Asy_b}), letting alone randomized algorithms. This confirms the asymptotically optimality of AdaWISH. Notice that the lower bound is proved assuming the optimal algorithm has access to the $ExactQuery$. Even given such a privilege, in the worst case the optimal algorithm has to make the number of queries that 
is in the same order of magnitude as that of AdaWISH assuming either neighboring or pointwise query oracles. Our proof does not depend on a specific querying scheme. The optimal algorithm is free to choose any scheme that minimizes the number of queries. 

Our proof sketch is as follows. Prove by contradiction. Suppose one algorithm $\mathcal{A}$ queries less than $(1-\epsilon)\frac{n}{\kappa^2}$ times and obtain a $\kappa$-approximation, then we can construct two functions $w_1(\sigma)$ and $w_2(\sigma)$, such that (i) both $w_1$ and $w_2$ match on all the queried points of algorithm $\mathcal{A}$, but (ii) the sum $W_1 = \sum_{\sigma} w_1(\sigma)$ and $W_2 =\sum_{\sigma} w_2(\sigma)$ differ more than a multiplicative factor of $\kappa^2$ (see Theorem~\ref{Th:Asy_b}). This suggests that algorithm $\mathcal{A}$ cannot give a $\kappa$-approximation, which contradicts with the assumption. 

\begin{Th}\label{Th:Asy_b}
 For $\kappa \geq 1$ and $0<\epsilon<1$, given any $(1-\epsilon)\frac{n}{\kappa^2}$ queried points on $w$, there always exists two functions $w_1(\sigma)$ and $w_2(\sigma)$ which match on the queried points, but  their corresponding sums $W_1$ and $W_2$ satisfy $W_2 > \kappa^2 W_1$.
\end{Th}
The proof to theorem~\ref{Th:Asy_b} is a careful mathematical construction, which is made available in \cite{fan2019towards}.
With theorem~\ref{Th:Asy_b}, we are able to show Corollary \ref{Co:Asy_b} which is the asymptotic lower bound.
\begin{Co}\label{Co:Asy_b}
\textbf{(Lower Bound)} Let $\mathcal{A}$ be any algorithm that accesses $w(\sigma)$ via the $\exactquery$ oracle. For any $\kappa\geq 1$ and $0<\epsilon<1$, $\mathcal{A}$ cannot guarantee a $\kappa$-approximation of $W=\sum_{\sigma} w(\sigma)$ for all $w$ if $A$ only issues $(1-\epsilon)\frac{n}{\kappa^2}$ $\exactquery$ calls.
\end{Co}
This shows that any algorithm that guarantees a $\kappa$-approximation for $\emph{all}$ $w$ must issue more than $(1-\epsilon) \frac{n}{k^2}$ queries on some input. Further, since this lower bound applies even to algorithms that have access to $\exactquery$, it also applies to weaker algorithms that only have access to $\approxquery$, and irrespective of whether they use a deterministic or randomized method to compute $\approxquery$.
%
We leave to future work an extension of this lower bound to the general case of randomized algorithms that guarantee a $\kappa$-approximation on only, say, $99\%$ of weight functions $w$.

Interestingly, the $T$ parameter in $\xorquery$ can also be reduced adaptively. In WISH, $T$ scales with $\log n$ because of the need for a union bound on estimation error introduced in each of the $n$ oracle queries. If AdaWISH makes only $k \leq n$ queries, then $T$ only needs to scale as $\log k$, thus reducing the amount of repetitions.

\subsection{Point-wise query oracle}

\begin{figure*}[ht]
\includegraphics[width=0.14\linewidth]{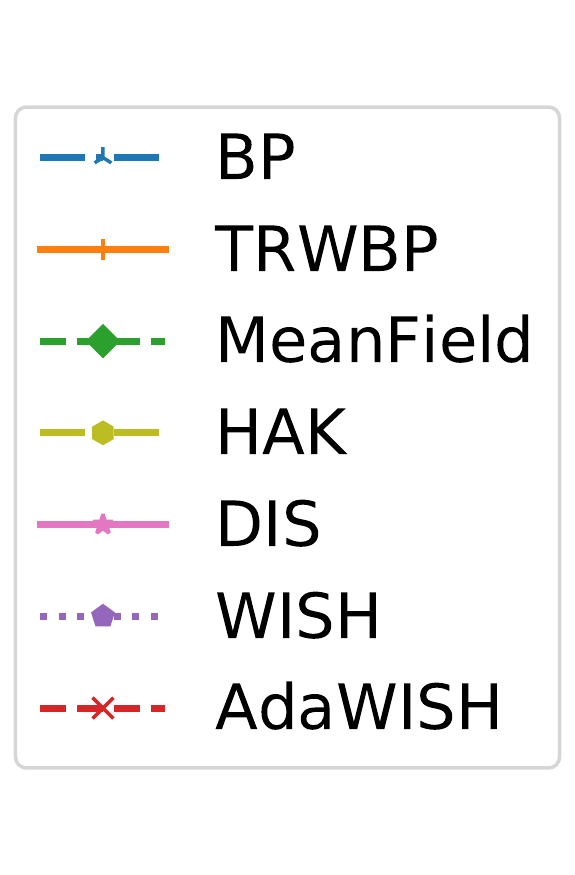}
\subfigure[Clique Ising]{\label{fig:cliq_est}
\includegraphics[width=0.29\linewidth]{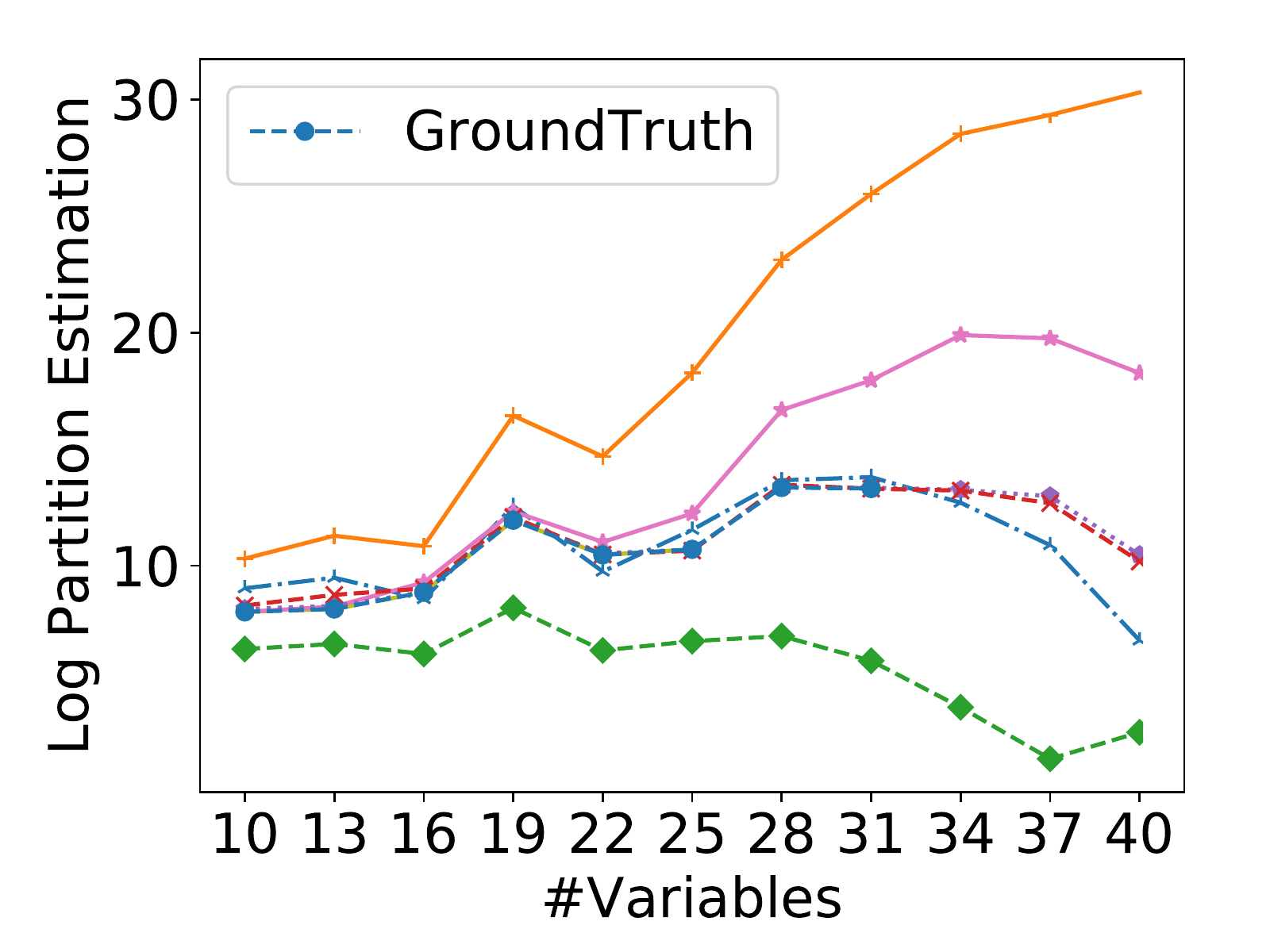}}
\hspace{-0.2in}
\subfigure[Mixed Grid Ising]{\label{fig:grid_mixed_est}
\includegraphics[width=0.29\linewidth]{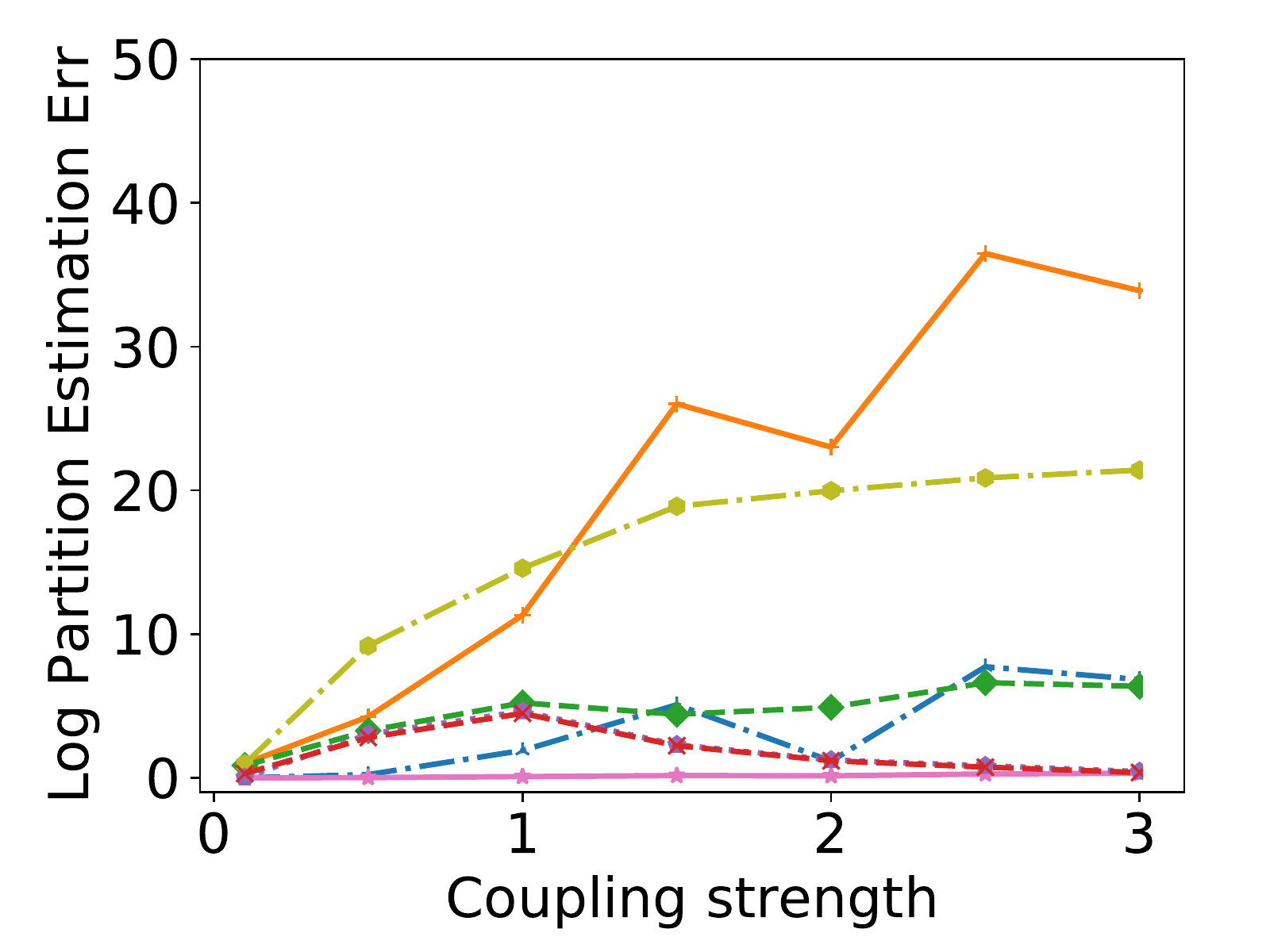}}
\hspace{-0.2in}
\subfigure[UAI inference]{\label{fig:UAIerror}
\includegraphics[width=0.29\linewidth]{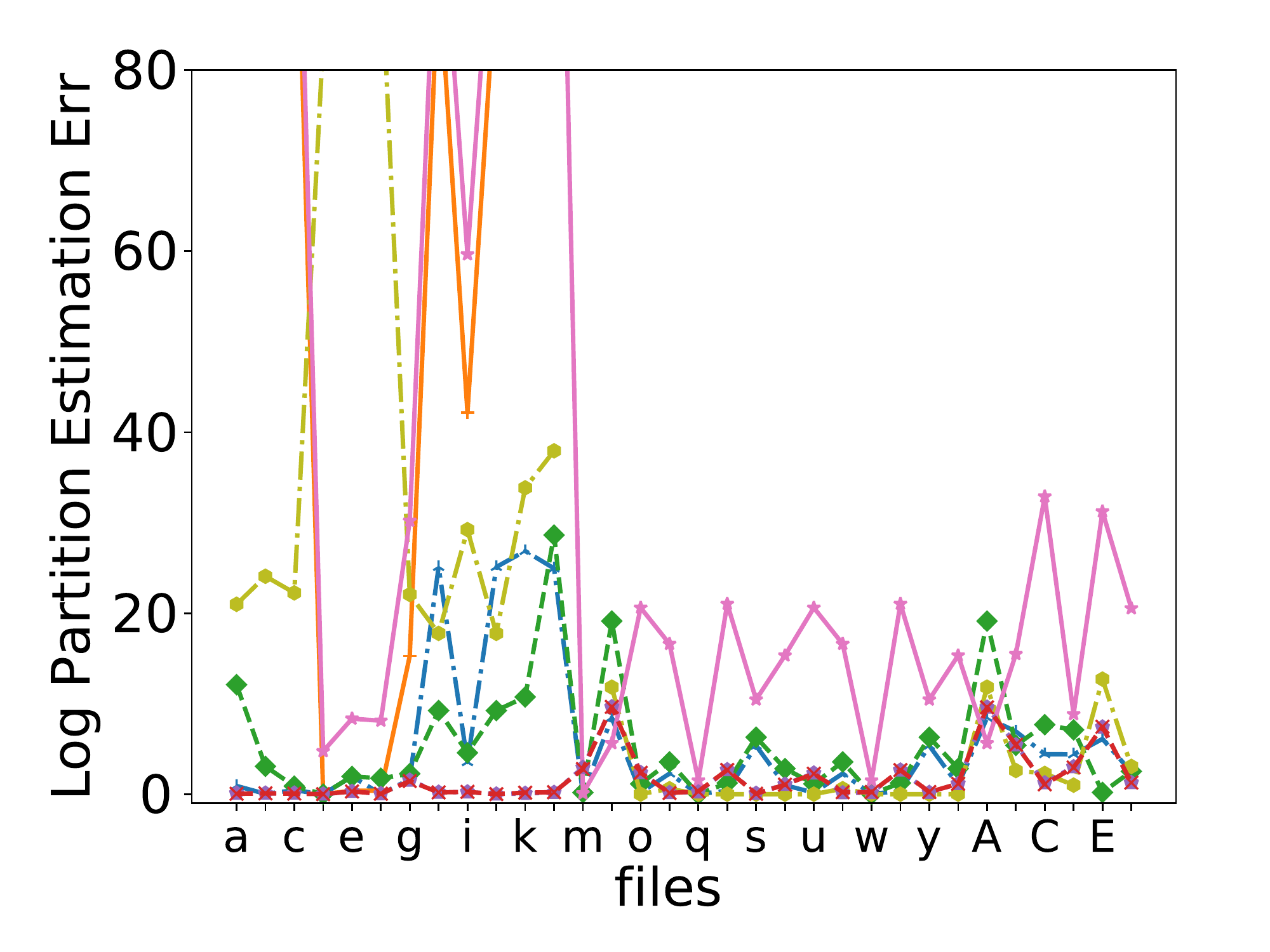}}

\includegraphics[width=0.14\linewidth]{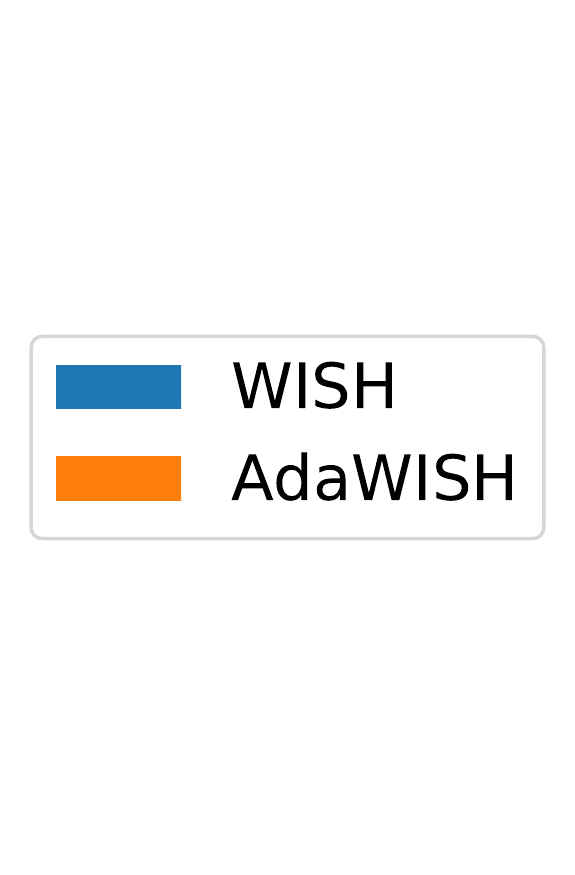}
\subfigure[Clique Ising]{\label{fig:cliquequery}
\includegraphics[width=0.29\linewidth]{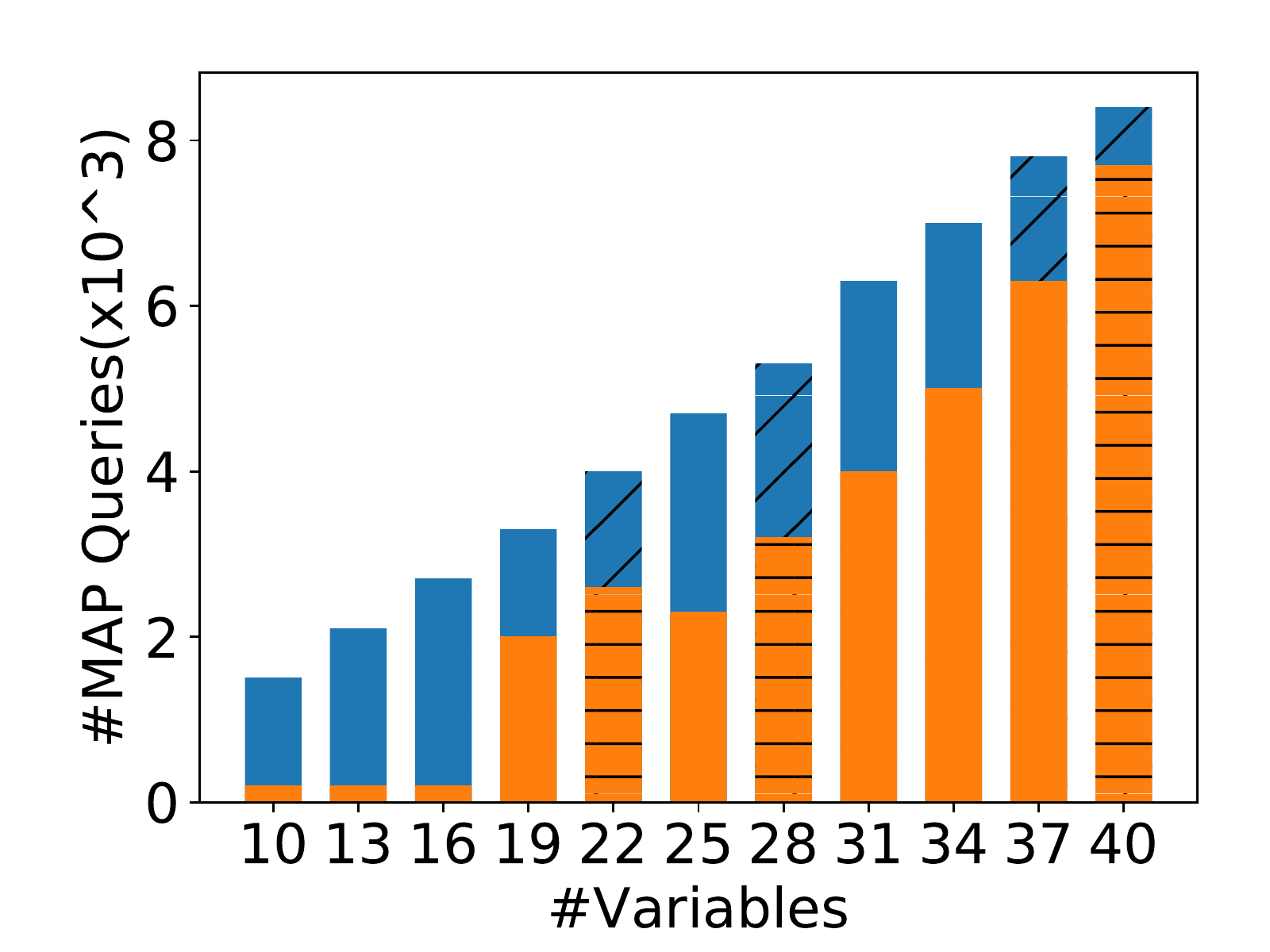}}
\hspace{-0.2in}
\subfigure[Mixed Grid Ising]{\label{fig:grid_mixed_qry}
\includegraphics[width=0.29\linewidth]{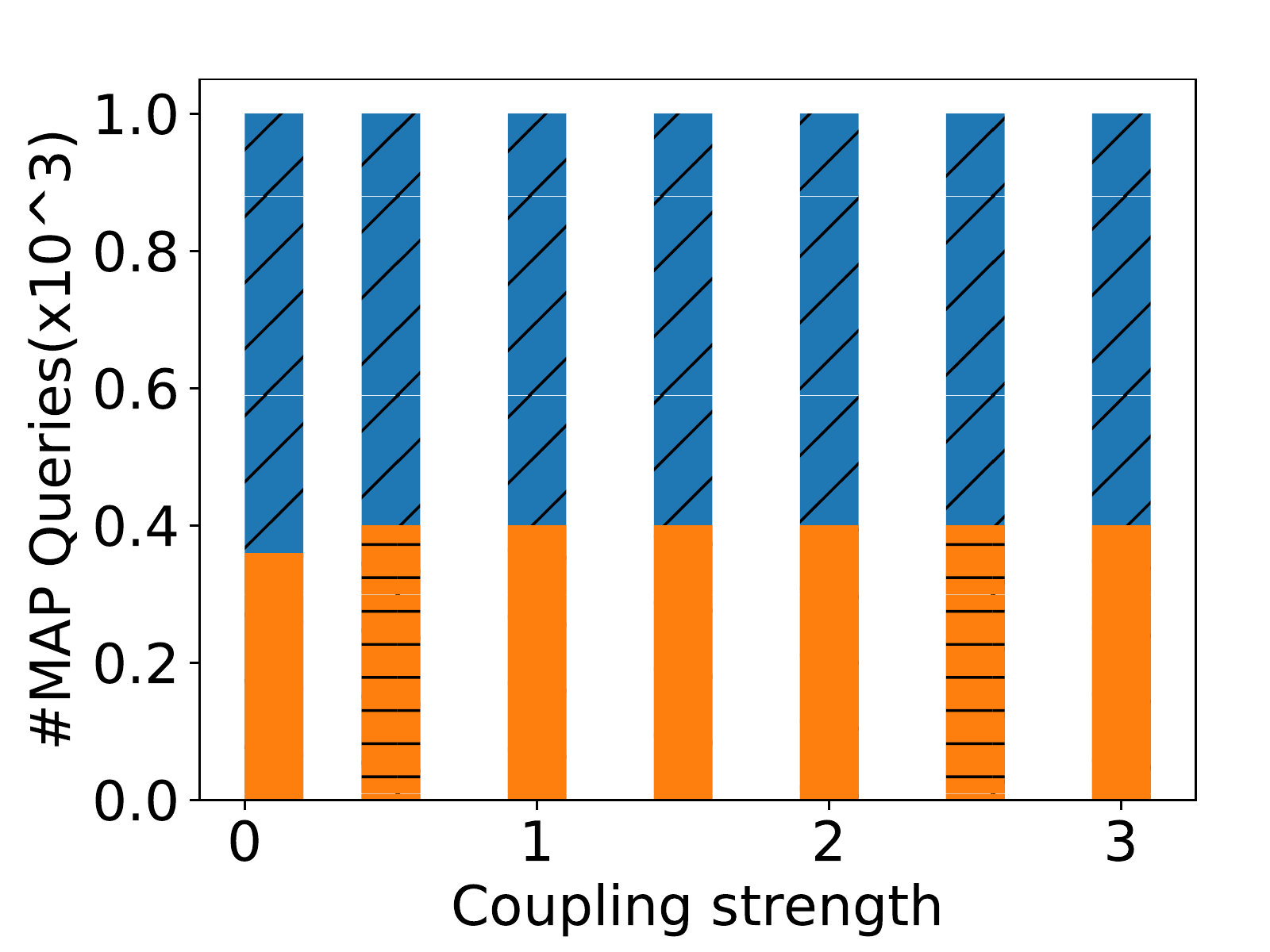}}
\hspace{-0.2in}
\subfigure[UAI inference]{\label{fig:UAIquery}
\includegraphics[width=0.29\linewidth]{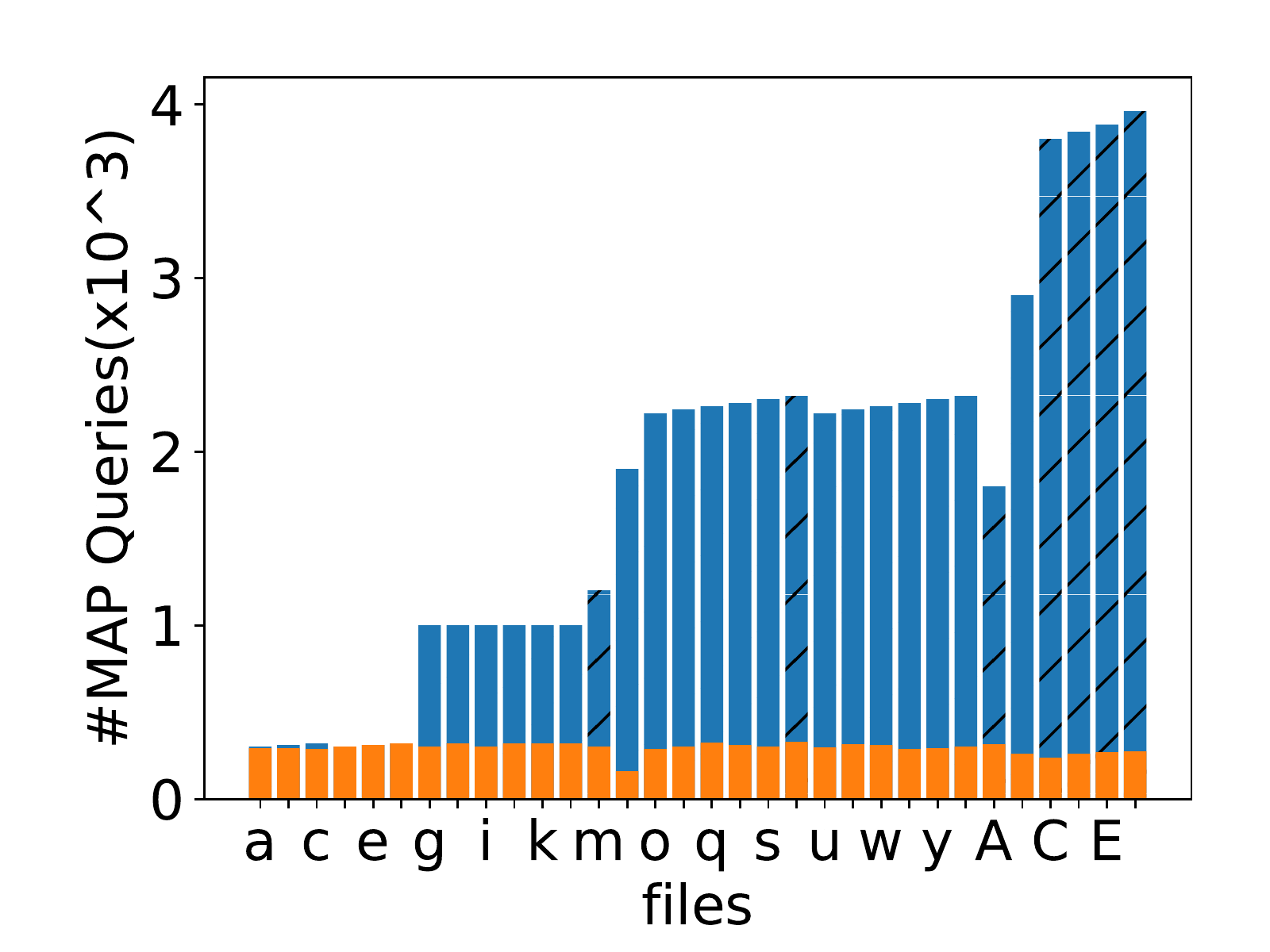}}
\caption{\textbf{(Top)} The errors in log partition function estimation for various methods on different benchmarks. \textbf{(Bottom)} The number of queries needed by WISH and AdaWISH. We can see that both WISH and AdaWISH provide good quality estimations. They are competetive with best benchmark methods on various tasks, much better than traditional methods such as MF, BP and TRWBP. AdaWISH needs much less number of queries than WISH (saving $47.7\%$ in median on Clique Ising models in the left column, $60\%$ of WISH queries on Mixed Grid Ising models in the middle column, and $81.5\%$ in median on UAI inference benchmarks in the right column)(bottom figures). 
}
\label{fig:ising_query}
\vspace{-0.1in}
\end{figure*}

Besides $\neighborquery$, we also consider point-wise approximation, which is also a natural relaxation.
We define $\pointquery$ as an oracle that has a pointwise approximation ratio $\gamma$: let $\tilde{b}_i$ be the returned value of $\pointquery(i, \Sigma, w)$. Then for all $i=0, \ldots, n$, we must have $b_i/\gamma \leq \tilde{b}_i \leq b_i \gamma$.
%
%
Given a pointwise query oracle, we are able to obtain a variant of AdaWISH which yields a $2\beta\gamma^2$-approximation of $W$. 
In our implementation, $\approxquery(i, \Sigma, w)$ returns exactly $\pointquery(i, \Sigma, w)$, 
$\upperbound(i, \Sigma, w)$ returns $\gamma \pointquery(i, \Sigma, w) $, and $\lowerbound(i, \Sigma, w)$ returns $\pointquery(i, \Sigma, w) / \gamma$. 
Notice that we also implement a lookup table within $\pointquery$ in this case, 
so the same query is not computed more than once. 
\begin{Th}\label{th:2beta}
Let $w,\Sigma,\gamma$ be as defined earlier. For any $\kappa > 2\gamma^2$, the output of AdaWISH (Algorithm \ref{alg:AdaWISHXOR}) on input ($\Sigma, w, \kappa / (2\gamma^2)$), assuming point-wise query oracle, is a $\kappa$-approximation of $W$.
\end{Th}

\noindent\textit{(Proof sketch)} We already know from Lemma \ref{Lm:wish} that there is a lower bound $LB=b_0+\sum_{i=0}^{n-1}b_{i+1}(2^{i+1}-2^i)$ and a upper bound $UB=b_0+\sum_{i=0}^{n-1}b_{i}(2^{i+1}-2^i)$ of $W$ which satisfy $LB\leq W\leq UB\leq 2LB$ if we query all $b_i$. That is to say, $LB$ is a 2-approximation of $W$. Since here we don't query all quantiles, we can find a relaxed lower bound $LB'$ and an upper bound $UB'$ of $W$. We construct $LB'$ and $UB'$ by replacing each $b_i$ in $LB$ and $UB$ with other values. For each $b_i$ returned from $\approxquery$, we use its returned value divided by $\gamma$ (times $\gamma$) as its new value in $LB'$ ($UB'$). 
In this case, the replacements of $b_i$ in $LB'$ and $UB'$ differ by $\gamma^2$. Because $\frac{\kappa}{2} > \gamma^2$ (condition of theorem~\ref{th:2beta}), the difference is less than $\frac{\kappa}{2}$. 
Otherwise, $b_i$ is contained in an interval satisfying stopping condition 2. We use the output of $\lowerbound$ on $b_i$'s immediate righthand side as its replacement in forming $LB'$; and use the output of  $\upperbound$ on $b_i$'s immediate lefthand side as its replacement in forming $UB'$. 
In this case, the difference of the replacements of $b_i$ in $LB'$ and $UB'$ is given by $\beta\gamma^2$ (stopping condition 2), which is $\beta\gamma^2 = \frac{\kappa}{2\gamma^2} \gamma^2 = \frac{\kappa}{2}$. 
Originally, $LB \leq UB \leq 2LB$ (Lemma~\ref{Lm:wish}). The replacement of each quantile $b_i$ in $LB'$ and $UB'$ further
broaden the distance by another factor of $\frac{\kappa}{2}$. 
Therefore, now $LB' \leq UB' \leq \frac{\kappa}{2} 2LB' = \kappa LB'$. 
Since $LB'$ and $UB'$ are still relaxed bounds, we must have $LB'\leq W \leq UB' \leq  \kappa LB'$. This concludes the proof. 

AdaWISH also satisfies the upper bound, the regret bound, and the asymptotic lower bound assuming the point-wise query oracle. %
In terms of the upper bound, it is worth noting that we also use a look-up table to save all the previous queried points as for $\neighborquery$. Therefore, our algorithm queries no more points than WISH.
We can also analyze the regret bound in a similar way as that assuming the $\neighborquery$. Our definition of the optimal algorithm is the same, which accesses an $ExactQuery$ oracle, and we can also prove a regret bound of $O(\log n)$ of $OPT$. 
%
In terms of asymptotic lower bound, Corollary \ref{Co:Asy_b} states that even the optimal algorithm, which has access to $ExactQuery$, has to query $\Omega(n)$ times in the worst case for a constant-approximation. 
AdaWISH, in this case, accesses the pointwise approximate oracle the same order of magnititude of times as the optimal algorithm accessing the $ExactQuery$ oracle, confirming its optimality.

\section{EXPERIMENTS}

In our experiments, we mainly focus on $NeighborQuery$ oracles which uses $\xorquery$ technique (algorithm \ref{alg:queryXOR}). We implement AdaWISH using IBM ILOG CPLEX Optimizer 12.9 for MAP queries.
We adopt the implementation of WISH \cite{Ermon13Wish} augmented with randomized low-density parity constraints \cite{Ermon14LowDensityParity}.
For comparison, we consider Junction Tree (JT), which provides exact inference results as ground truth, Tree-Reweighted Belief propagation (TRWBP) \cite{WainwrightJW03}, which provides an upper bound on the partition function, Mean Field (MF) \cite{Wainwright2008}, which provides a lower bound, Loopy Belief Propagation (BP) \cite{murphy1999loopy}, which has no guarantees, and Double-loop GBP (HAK) \cite{Heskes2002HAK}, which is a winning solver in the UAI inference challenge. We use the implementations of these algorithms available in LibDAI \cite{Mooij2010libDAI}. 
We also compare with Dynamic Importance Sampling (DIS) \cite{lou2017dynamic,lou2019interleave}, a recently proposed strong baseline. 
While we use JT here for ground truth, ACE is a faster alternative for exact inference and, in our tests, provided estimates that agree with JT up to the internal precision.
%
%
The inference problem we consider is to compute the partition function.

%

Our main result is shown in Figure~\ref{fig:ising_query}. In a nutshell, on all three sets of benchmarks, AdaWISH gives precise estimation to the partition function with estimated curves overlapping with those of WISH algorithm, better than competing approaches. Meanwhile, AdaWISH saves plenty of queries.
%





%
\begin{table}
\centering
\caption{The benchmarks from UAI 2011 and 2014 inference competition used in the evaluation in Figure \ref{fig:UAIerror} and \ref{fig:UAIquery}. 
The first (third) column shows the abbreviations used in Figure \ref{fig:UAIerror} and \ref{fig:UAIquery} and the second (fourth) shows the names of the benchmarks. 
}
\begin{tabular}{|c|l||c|l|}
\hline
a & rbm\_40 & q & Segmentation\_229 \\
b & rbm\_42 & r & Segmentation\_231 \\
c & rbm\_44 & s & Segmentation\_232 \\
d & rbm\_ferro\_40 & t & Segmentation\_235 \\
e & rbm\_ferro\_42 & u & 226binary \\
f & rbm\_ferro\_44 & v & 228binary \\
g & Grids100f0 & w & 229binary \\
h & Grids100f1 & x & 231binary \\
i & Grids100f2 & y & 232binary \\
j & Grids100f3 & z & 235binary \\
k & Grids100f4 & A & Promedus\_200 \\
l & Grids100f5 & B & Promedus\_306 \\
m & smokers\_120 & C & Promedus\_374 \\
n & or\_chain\_200 & D & Promedus\_378 \\
o & Segmentation\_226 & E & Promedus\_385 \\
p & Segmentation\_228 & F & Promedus\_400 \\
\hline
\end{tabular}
\label{tab:uai}
\end{table}
~\\
\noindent\textbf{Clique Ising model.} ~We first consider random Clique-structured Ising models with $n$ binary variables $x_i\in\{0,1\}$ for $i \in \{1,..,n\}$. %
Let $\sigma= (x_1, \ldots, x_n)$. 
The Ising model is to sum over $w(\sigma) = \exp(-\sum_{i,j} w_{ij} x_i x_j)$, where $w_{ij}=0$ if $i=j$, otherwise uniformly sampled from $[0,w\sqrt{|i-j|}]$. 
The coupling strength $w$ is set to be $0.1$.
%
We introduce two closed chains of strong repulsive interactions with a length of $0.3n$. The strengths of interactions are uniformly sampled from $[0, 100w]$. Note that Clique Ising models have treewidth $n$ and therefore those with more than $31$ variables can not be solved exactly (The curve for ground truth terminates at the case of 31 variables in Figure~\ref{fig:cliq_est}). 
In our experiments, each MAP query 
is carried out using CPLEX on a single core,
and experiment is carried out on a cluster, where each node has 24 cores and 96GB memory. We follow the design that each $\approxquery$ issued by AdaWISH consists of a batch of MAP queries that can be executed in parallel. $\approxquery$ with the same binary search depth can be issued and executed simultaneously. The time of AdaWISH is therefore the sum of maximum time taken by a single MAP query of each binary search depth. 

In this experiment, we run both WISH and AdaWISH to give results with provable guarantees. 
Let $\beta, c, \delta$ as defined before, then both of WISH and AdaWISH use the settings of $c=5, \delta=0.01, \alpha(c)=0.078$, where $\alpha$ is a function of $c$ as in \cite{Ermon13Wish}. And the trade-off parameter is set to be $\beta=10^5$ for AdaWISH. In this benchmark, we set $T=\frac{\ln{1/\delta}}{\alpha}\ln{n}$ and execute AdaWISH/WISH to optimality to get results with theoretical guarantees based on Theorem \ref{th:2beta2c}. As a fair comparison, all competing approaches are also executed to their optimality. 
With the current parameter setting, we guarantee that with the probability at least $(1-\delta)$ the error of AdaWISH/WISH is no larger than 8 in $\log{10}$ scale.

Figure \ref{fig:cliq_est} reports the performance of all the methods on generated random Clique Ising models. The theoretical guarantee provided by Theorem \ref{th:2beta2c} holds since all  optimizations are solved to optimality. Note that it plots their estimations rather than errors since we can only get the ground truth for models with less than $31$ variables. We can see that the estimations given by WISH and AdaWISH are visually overlapping with the plots of ground truth, outperforming the second best method BP in stability. As for other methods, TRWBP and MF diverge from the ground truth. HAK fails to return valid results therefore is absent
 in the graph. We also note that the empirical result of AdaWISH is much better than the worst-case theoretical guarantee. AdaWISH also saves $47.7\%$ in median on the number of queries when compared to WISH (figure \ref{fig:cliquequery}). 
 

~\\
\noindent\textbf{Grid Ising model.} ~We next investigate the performance of AdaWISH on
large instances. 
In the following two experiments, we focus on 
the empirical performance, therefore relaxing the theoretic guarantees, which are extremely expensive to obtain.
Note only WISH/AdaWISH is able to obtain constant approximation guarantees among all approaches in
the experiment. 
%
%
Let $\sigma = (x_1,x_2...x_n) \in\{-1,1\}^n$. 
The grid Ising model is to sum over $w(\sigma)= \exp(\sum_i f_i x_i + \sum_{(i,j) \in \mbox{grid}} w_{ij} x_i x_j)$.
Here, $f_i$ is sampled uniformly from $[-0.1, 0.1]$, $w_{ij}$ is sampled from $[-w, w]$, where $w$ is called the coupling strength, and is shown in the horizontal axis of Figure~\ref{fig:grid_mixed_est}. 
We insert structures in the grids by introducing a rectangle of strong interactions, inside which the coupling strength is amplified by 10.

Both of WISH and AdaWISH use the settings of $c=5, T=10$, and AdaWISH uses $\beta=100$. 
In these two experiments, we set a timeout of 
4 hours for all algorithms and force each MAP query issued by WISH/AdaWISH to stop in 15 minutes, so not all the query oracles of WISH/AdaWISH are solved up to optimality. %
For each coupling strength we generate $10$ instances and report the median error in the log10 partition function estimation.

From Figure~\ref{fig:grid_mixed_est}, we can see that AdaWISH and WISH are competitive with the state-of-the-art method DIS, outperform other methods. DIS works well on Ising models, but not well on UAI instances as we show in the next section. 
When coupling strength is larger than $2.5$, WISH and AdaWISH can give near-optimal estimations, with an error of roughly $0.8$ in the log10 partition function, while a strong competing approach, BP, has an log10 error of $6$. 
Meanwhile, in Figure \ref{fig:grid_mixed_qry}, AdaWISH reduces approximately $60\%$ queries from WISH, while giving the same or even better estimations.
%

~\\
\noindent\textbf{UAI inference.} ~We also test AdaWISH on open datasets in the UAI Approximate Inference Challenge.\footnote{http://auai.org/uai2014/competition.shtml} The same settings of $c,\beta$ and T are adopted as those for grid Ising models but the timeout is increased to 30 minutes for each MAP query due to the difficulty of the problems. Figure \ref{fig:UAIerror} reports the estimation error of the log10-partition function for various methods. 
Inference benchmark names corresponding to the abbreviations used in Figures~\ref{fig:UAIerror} and \ref{fig:UAIquery} can be found in Table \ref{tab:uai}. While we used all instances, due to limited space on the x-axis, we only labelled every other instance in Figures~\ref{fig:UAIerror} and \ref{fig:UAIquery}.

Here, we see that AdaWISH and WISH have small errors in estimating the partition functions of all instances, outperforming other methods in accuracy. 
A few points are missing for several approaches (for example, TRWBP and HAK) because their errors are too big or failing to complete within the time given. 
%
%
Meanwhile, as shown in Figure \ref{fig:UAIquery}, the number of queries of AdaWISH is stable at around $400$  over different benchmarks, saving $81.5\%$ queries in median when compared to WISH.



\section{CONCLUSION}

We introduced AdaWISH, an algorithm based on hashing and optimization that provides a constant-factor approximation of the discrete integration problem. %
AdaWISH significantly reduces the number of optimization queries needed by WISH via an adaptive  binary search, while continuing to provide equally precise results for discrete integration.
%
%
The number of queries made by AdaWISH is, in fact, provably no more than a logarithmic factor of that of an optimal  algorithm that knows the shape of the function apriori.
In addition, the number of optimization queries made by AdaWISH is asymptotically optimal because even the optimal algorithm has to make the same order of magnititude number of queries as AdaWISH in the worst case.
We evaluate the performance of AdaWISH on a collection of challenging benchmarks. Empirically, AdaWISH gives precise estimations, of the same quality as that of WISH, significantly outperforming competing approaches. Meanwhile, AdaWISH issues 47\%-60\% fewer queries on Ising models, and saves in median 81.5\% of the queries on benchmarks in the UAI inference challenge. 


\section{ACKNOWLEDGEMENTS}
 This research was supported by the National Science Foundation (Award number IIS-1850243 and CCF-1918327). The computing infrastructure was partially supported by the Microsoft AI for Earth computing award. The authors also acknowledge the support from the Allen Institute of AI.

\begin{small}
\bibliographystyle{ecai}
\bibliography{xue}
\end{small}


\clearpage
\appendix

\section{SUPPLEMENTARY MATERIAL}


\subsection{Proofs}
\label{appendix:proof}

\subsubsection{Proof of Lemma \ref{Lm:wish}}

\begin{proof} (Lemma~\ref{Lm:wish})
We have $b_i \geq w(\sigma_k) \geq b_{i+1}$ for all $k \in \{2^i, \ldots, 2^{i+1}\}$. Therefore, a lower bound can be obtained by replacing $w(\sigma_k)$ with $b_{i+1}$ in the summation, while an upper bound can be obtained by replacing each $w(\sigma_k)$ with $b_{i}$. This yields a lower bound of $\mathit{LB} = b_0 + \sum_{i=1}^n b_i (2^i - 2^{i-1})$ and an upper bound of $\mathit{UB} = b_0 + \sum_{i=1}^n b_{i-1} (2^i - 2^{i-1})$. That $\mathit{LB} \leq W \leq \mathit{UB}$ follows from the construction. Further, we have:
\begin{align*}
\mathit{UB} &=  b_0 + \sum_{i=1}^n b_{i-1} (2^i - 2^{i-1})\\
       &=b_0 + b_0 + \sum_{i=1}^{n-1} 2b_{i} (2^i - 2^{i-1})\\
       &\leq 2b_0+\sum_{i=1}^{n-1} 2b_{i} (2^i - 2^{i-1})\\
       &\leq 2(b_0+\sum_{i=1}^{n-1} b_i (2^i - 2^{i-1}))\\
       &\leq 2 \mathit{LB}
\end{align*}
This finishes the proof.
\end{proof}

\subsubsection{Proof of Theorem \ref{Th:Asy_b}}\label{app:lowerbound}

\begin{proof}(Theorem \ref{Th:Asy_b})
Our proof works by constructing two weight functions $w_1(\sigma)$ and $w_2(\sigma)$ passing through the shared $(1-\epsilon)\frac{n}{\kappa^2}$ queried points such that their discrete integrals $W_1$ and $W_2$, respectively, are a factor of at least $\kappa^2$ apart. To achieve this, $w_1(\sigma)$ and $w_2(\sigma)$ are constructed such that $w(\sigma_j)=b_{r}$ in $w_1(\sigma)$ and $w(\sigma_j)=b_{l}$ in $w_2(\sigma)$ for all $j\in\{2^l,2^{r}\}$, where $b_l$ and $b_r$ are the neighbor queried points among those $(1-\epsilon)\frac{n}{\kappa^2}$ points. If we query all of these $n+1$ points, $W_1$ is $b_0 + \sum_{i=0}^{n-1} b_{i+1} (2^{i+1} - 2^i)$ and $W_2$ is $b_0 + \sum_{i=0}^{n-1} b_i (2^{i+1} - 2^i)$. 

First consider the case of $\epsilon=0$. Here we construct a special series of $b_i$, namely $b_i=\frac{1}{2^i}$, such that each interval is of the same importance in terms of its contribution to the integral of the weight function. Intuitively, this means we should keep the $\frac{n}{\kappa^2}$ queried points as scattered as possible. We accomplish this by querying every other $\kappa^2$ apart point. This makes $W_1$ the largest and $W_2$ the smallest possible, leading to the least possible $\frac{W_2}{W_1}$. We claim that even in this situation, $\frac{W_2}{W_1}$ is still larger than $\kappa^2$.

Firstly, we simplify the expression for $W_1$ to obtain:
\begin{align*}
W_1 &=1+\sum_{i=0}^{n/\kappa^2}((\kappa^2)^{i+1}-(\kappa^2)^i)\frac{1}{(\kappa^2)^{i+1}}\\
    &=1+\frac{n}{\kappa^2}(1-\frac{1}{\kappa^2})
\end{align*}
Meanwhile, $W_2$ can be written as
\begin{align*}
W_2&=1+\sum_{i=0}^{n/\kappa^2}((\kappa^2)^{i+1}-(\kappa^2)^i)\frac{1}{(\kappa^2)^{i}}\\
    &=1+\frac{n}{\kappa^2}(\kappa^2-1)
\end{align*}

Therefore, we can calculate the limit of $\frac{W_2}{W_1}$ as $n$ tends to infinity:
\begin{align*}
\lim_{n \rightarrow \infty}\frac{W_2}{W_1} &= \kappa^2
\end{align*}

Returning to the case of $\epsilon$ being a fixed, non-zero constant, we make the following observation: if we do not omit $\epsilon$ in the above calculation, $W_2$ would increase by $O(n)$ and $W_1$ would decrease by $O(n)$, leading to:
\begin{align*}
    \frac{W_2}{W_1} > \kappa^2
\end{align*}
 Therefore, $W_2>\kappa^2 W_1$ holds if we only allow $(1-\epsilon)(\frac{n}{\kappa^2})$ accesses to QUERY. This finishes the proof.
\end{proof}

\subsubsection{Proof of Corollary \ref{Co:Asy_b}}
\begin{proof} (Corollary \ref{Co:Asy_b})
Consider any algorithm $\mathcal{A}$ with only $(1-\epsilon)\frac{n}{\kappa^2}$ accesses to $ExactQuery$, which outputs $\tilde{W}$. Given these queried points, Theorem \ref{Th:Asy_b} implies that there exist two function $w_1(\sigma)$ and $w_2(\sigma)$ matching on these queried points with corresponding discrete integral $W_1$ and $W_2$ where $W_2 > \kappa^2 W_1$. Therefore, algorithm $\mathcal{A}$ output $\tilde{W}$ as the estimate for both $W_1$ and $W_2$.

Assume, for the sake of contradiction, that $\tilde{W}$ is a $\kappa$-approximation of both $W_1$ and $W_2$. Then we have:
$$\frac{\tilde{W}}{\kappa} \leq W_1 \leq \kappa\tilde{W}$$
and 
$$\frac{\tilde{W}}{\kappa} \leq W_2 \leq \kappa\tilde{W}$$
Together, these imply $W_2\leq \kappa^2W_1$, which contradicts the above conclusion $W_2 > \kappa^2 W_1$ from Theorem \ref{Th:Asy_b}. Hence, $\tilde{W}$ could not have been a $\kappa$-approximation of at least one of $W_1$ and $W_2$.
\end{proof}

\end{document}